\pdfoutput=1

\documentclass{article}

\usepackage{microtype}
\usepackage{graphicx}
\usepackage{booktabs} 
\usepackage{epsfig}
\usepackage{amsmath}
\usepackage{mathtools}
\usepackage{tabularx}
\usepackage{multirow}

\usepackage{hyperref}

\usepackage[accepted]{icml2019}

\icmltitlerunning{Improving Adversarial Robustness of Ensembles with Diversity Training}
\begin{document}

\twocolumn[
\icmltitle{Improving Adversarial Robustness of Ensembles with Diversity Training}



\begin{icmlauthorlist}

\icmlauthor{Sanjay Kariyappa}{gt}
\icmlauthor{Moinuddin K. Qureshi}{gt}
\end{icmlauthorlist}

\icmlaffiliation{gt}{Georgia Institute of Technology}

\icmlcorrespondingauthor{Sanjay Kariyappa}{sanjaykariyappa@gatech.edu}
\icmlcorrespondingauthor{Moinuddin K. Qureshi}{moin@gatech.edu}

\icmlkeywords{Machine Learning, ICML}

\vskip 0.3in
]



\printAffiliationsAndNotice{} 
\newcommand{\ignore}[1]{}

\ignore{
This paper proposes Diversity Training, a novel method to reduce the transferability of adversarial examples in ensembles and better defend against black-box attacks. We identify that attacking ensembles requires the attack vector to lie in the shared adversarial subspace of at least half of the models in the ensemble. By using a new regularization term called Jacobain Alignment loss we show that the dimensionality of the shared adversarial subspace can be reduced, improving adversarial robustness. Our method can be used in conjunction with existing defenses to achieve improved adversarial robustness.
}

\ignore{Acutal abstract
Deep Neural Networks are vulnerable to adversarial attacks even in settings where the attacker has no direct access to the model being attacked. Such attacks usually rely on the principal of {\em transferability}, whereby an attack crafted on a surrogate model tends to transfer to the target model. We show that an ensemble of {\em diverse} models can provide an effective defense against this class of attacks. Our key insight is that an adversarial example is less likely to fool multiple models in the ensemble if the models have uncorrelated loss functions. To this end, we propose a new defense called {\em Diversity Training} which uses a novel regularization function called {\em Gradient Alignment Loss} 
to train an ensemble of diverse models with uncorrelated loss functions. Our method can be used in conjunction with existing defenses to significantly improve the adversarial robustness of ensembles.
}

\ignore{
This paper proposes Diversity Training, a novel method to improve the adversarial robustness of ensembles against black-box attacks. Attacking an ensemble requires the attack vector to fool at least half of the models in the ensemble. Our key insight is that it would be harder to craft adversarial perturbations against an ensemble of {\em diverse} models with dissimilar loss landscapes. We propose a new form of regularization called {\em Gradient Alignment Loss}, which can be used to train an ensemble of diverse models. Our method can be used in conjunction with existing defenses to significantly improve the adversarial robustness of ensembles.
}
\ignore{ # submission text
Deep Neural Networks are vulnerable to adversarial attacks even in settings where the attacker has no direct access to the model being attacked. Such attacks usually rely on the principal of transferability, whereby an attack crafted on a surrogate model tends to transfer to the target model. We show that an ensemble of diverse models can provide an effective defense against this class of attacks. Our key insight is that an adversarial example is less likely to fool multiple models in the ensemble if the models have uncorrelated loss functions. To this end, we propose a new defense called Diversity Training which uses a novel regularization function called Gradient Alignment Loss to train an ensemble of diverse models with uncorrelated loss functions. Our method can be used in conjunction with existing defenses to significantly improve the adversarial robustness of ensembles.
}
\begin{abstract}
Deep Neural Networks are vulnerable to adversarial attacks even in settings where the attacker has no direct access to the model being attacked. Such attacks usually rely on the principle of {\em transferability}, whereby an attack crafted on a surrogate model tends to transfer to the target model. We show that an ensemble of models with misaligned loss gradients can provide an effective defense against transfer-based attacks. Our key insight is that an adversarial example is less likely to fool multiple models in the ensemble if their loss functions do not increase in a correlated fashion. To this end, we propose {\em Diversity Training}, a novel method to train an ensemble of models with uncorrelated loss functions. We show that our method significantly improves the adversarial robustness of ensembles and can also be combined with existing methods to create a stronger defense.
\end{abstract}

\ignore{ submission
Deep Neural Networks are vulnerable to adversarial attacks even in settings where the attacker has no direct access to the model being attacked. Such attacks usually rely on the principle of transferability, whereby an attack crafted on a surrogate model tends to transfer to the target model. We show that an ensemble of models with misaligned loss gradients can provide an effective defense against transfer-based attacks. Our key insight is that an adversarial example is less likely to fool multiple models in the ensemble if their loss functions do not increase in a correlated fashion. To this end, we propose Diversity Training, a novel method to train an ensemble of models with uncorrelated loss functions. We show that our method significantly improves the adversarial robustness of ensembles and can also be combined with existing methods to create a stronger defense.
}
\section{Introduction}
\ignore{
1. What is Adv ML, why is it important (problem)
2. transferability (fig?) (problem) -- subsection
3. Ensembles as defense (Fig 1) (insight/goal) --subsection
4. Our Contribution
}
Despite achieving state of the art classification accuracies on a wide variety of tasks, it has been demonstrated that deep neural networks can be fooled into misclassifying an input that has been adversarially perturbed \cite{intrigue, exp_harness}. These adversarial perturbations are small enough to go unnoticed by humans but can reliably fool deep neural networks. The existence of adversarial inputs presents a security vulnerability in the deployment of deep neural networks for real world applications such as self-driving cars, online content moderation and malware detection. It is important to ensure that the models used in these applications are robust to adversarial inputs as failure to do so can have severe consequences ranging from loss in revenue to loss of lives. 

A number of attacks have been proposed which use the gradient information of the model to figure out the perturbations to a benign input that would make it adversarial. These attacks require access to the model parameters and are termed {\em white-box} attacks. Fortunately, several real world applications of deep learning do not expose the model parameters to the end user, making it harder for an adversary to attack the model. However, adversarial examples have been shown to {\em transfer} across different models \cite{tfr_ml}, enabling adversarial attacks without knowledge of the model architecture or parameters. 
Attacks that work under these constraints are termed {\em black-box attacks}. 

\subsection{Transferability}
Black-box attacks rely on the principle of {\em transferability}. In the absence of access to the target model, the adversary trains a surrogate model and crafts adversarial attacks on this model using white-box attacks. Adversarial examples generated this way can be used to fool the target model with a high probability of success ~\cite{transferable_adv_examples}. 
Adversarial examples have been known to span a large contiguous subspace of the input ~\cite{exp_harness}. Furthermore, recent work explaining transferability ~\cite{space} has shown that models with high dimensionality of adversarial subspace (Adv-SS) are more likely to intersect, causing adversarial examples to transfer between models.  Our goal is to find an effective way to reduce the dimensionality of the Adv-SS and hence reduce transferability of adversarial examples. We show that this can be done by using an ensemble of diversely trained models.

\begin{figure*}[htb]
	\centering
    \centerline{\epsfig{file=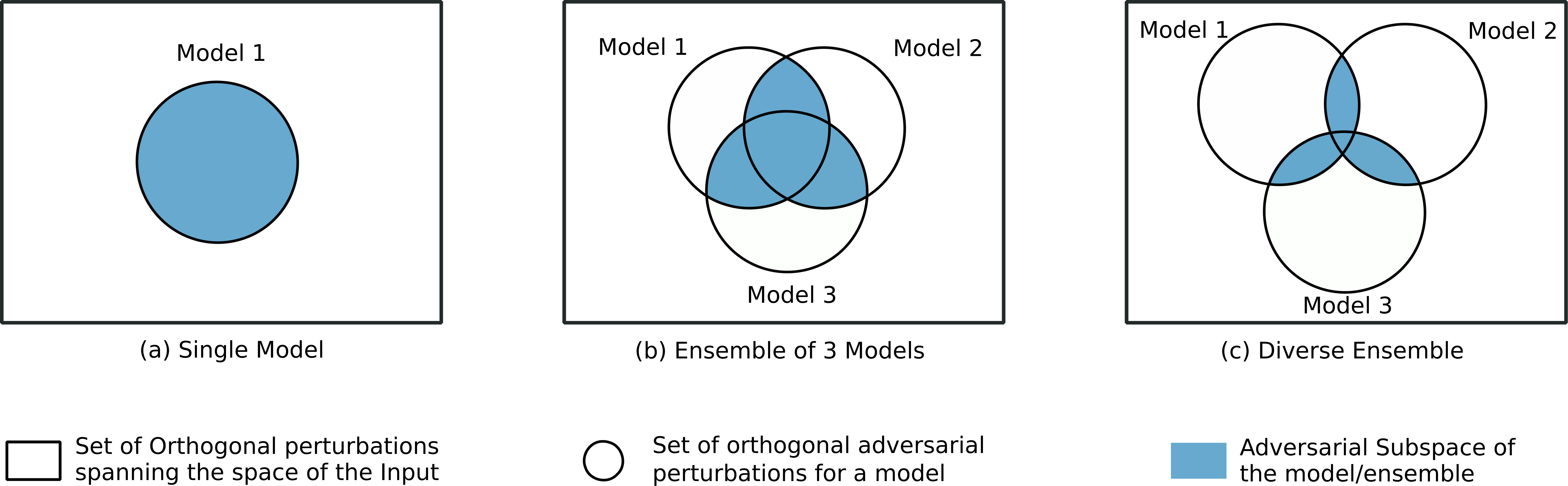, height=1.8in}}
	\caption{Venn diagram illustrations of the adversarial subspace of (a) single model (b) Ensemble of 3 models and (c) Diverse Ensemble. Our goal is to reduce the overlap in the adversarial subspaces of the models in the ensemble as shown in (c) }
	\vspace{0.1 in}
    \label{fig:motivation}
\end{figure*}

\subsection{Ensemble as an Effective Defense} \label{sec:ens_effective_defense}
An ensemble uses a collection of different models and aggregates their outputs for classification. Adversarial robustness can potentially be improved using an ensemble, as the attack vector must now fool multiple models in the ensemble instead of just a single model. In other words, the attack vector must now lie in the shared Adv-SS of the models in the ensemble. We explain this using the Venn diagrams in  Figure~\ref{fig:motivation}. The area enclosed by the rectangle represents a set of orthogonal perturbations spanning the space of the input and the circle represents the subset of these orthogonal perturbations that are adversarial (i.e. cause the model to misclassify the input). The shaded region represents the Adv-SS. For a single model (Figure~\ref{fig:motivation}a), any perturbation that lies in the subspace defined by the circle would cause the model to misclassify the input. In contrast, for an ensemble of 3 models, the adversarial input must fool more than one model in the ensemble for the attack to be successful. This implies that the attack vector must lie in the shared Adv-SS of the ensemble as shown in Figure~\ref{fig:motivation}b.

While prior works have used ensemble of models in various forms to defend against adversarial attacks ~\cite{random_self_ensemble,ensemble_defense}, we show that the efficacy of ensembles can be improved significantly by explicitly forcing a reduction in the shared Adv-SS of the models as shown in Figure~\ref{fig:motivation}c. Reducing this overlap translates to a reduction in the overall dimensionality of the Adv-SS of the ensemble. Thus, there are fewer directions of adversarial perturbations that can cause multiple models in the ensemble to misclassify, resulting in a reduction of transferability and improved adversarial robustness.

\subsection{Contributions}
We study the use of ensembles in the context of black-box attacks and propose a technique to improve adversarial robustness. Overall, we make the following key contributions:
\begin{enumerate}
\item We identify that the adversarial robustness of an ensemble can be improved by reducing the dimensionality of the shared adversarial subspace
\item We propose {\em Gradient Alignment Loss(GAL)}, a metric to measure the shared adversarial subspace of the models in an ensemble.
\item We show that GAL can be used as a regularizer to train an ensemble of {\em diverse} models with misaligned loss gradients. We call this {\em Diversity Training}.
\item We show empirically that Diversity Training makes ensembles more robust to transfer-based attacks.
\ignore{
\item A novel technique to improve the adversarial robustness of an ensemble called {\em Diversity Training} is proposed
\item We evaluate our defense using vgg-16 and Resnet-18 architectures for mnist and cifar-10 datasets
\item We validate our claims of reduced adversarial subspace using Gradient Aligned Adversarial Subspace Analysis
}
\end{enumerate}
\section{Background}
In this section we formally define the attack model and describe the various attacks considered in our work. 
\subsection{Adversarial Examples}
 A benign input $x$ can be transformed into an adversarial input $x'$ by adding a carefully crafted perturbation $\eta$
\begin{gather}
x' = x + \eta
\end{gather}
For an untargeted attack, the adversary's objective is to cause the model $f$ to misclassify the perturbed input such that $f(x') \neq y_{true}$, where $y_{true}$ is the ground-truth label of $x$.
However, this perturbation must not result in a perceivable change to the input for a human observer. Following prior work on adversarial machine learning for image classification~\cite{exp_harness,madry2018towards}, we enforce this constraint by restricting the $l_{\infty}$ norm of the perturbation to be below a threshold $\epsilon$ i.e. $\rVert\eta\rVert_{\infty} \leq \epsilon$.

\subsection{Black-Box Attack Model}
This work considers the black-box attack model in which the attacker does not know the parameters of the target model. We assume however that the adversary has access to the dataset used for training and knows the architecture of the model being attacked. 
To attack a target model $T$ trained on dataset $\mathcal{D}$, the adversary trains a surrogate model $S$ using the same dataset. 
Adversarial examples crafted on the surrogate model using white-box attacks can be used to attack the target model using the principle of transferability. 
We briefly describe the various attack algorithms considered in our evaluations that can be used for this purpose.

\subsection{Attack Algorithms} \label{ssec:wb_attacks}

Given full access to the model parameters, the adversary can craft an adversarial example by considering the loss function. Let $J(\theta,x,y)$ denote the loss function of the model $f$, where $\theta$ represents the model parameters, $x$ is the benign input and $y$ is the label. The attacker's goal is to generate an adversarial example $x'$ with the objective of maximizing the model's loss function, such that $J(\theta,x',y) > J(\theta,x,y)$,  while adhering to the constraint: $\rVert x'-x \rVert_{\infty} \leq \epsilon$. Several techniques have been proposed in literature to solve this constrained optimization problem. We discuss the ones used in the evaluation of our defense.

\textbf{Fast Gradient Sign Method (FGSM):} FGSM ~\cite{exp_harness} uses a linear approximation of the loss function to find an adversarial perturbation that causes the loss function to increase. Let $\nabla_{x} J(\theta,x,y)$ denote the gradient of the loss function with respect to $x$. The input is modified by adding a perturbation of size $\epsilon$ in the direction of the gradient vector.
\begin{gather} 
x' = x + \epsilon \cdot sign(\nabla_{x} J(\theta,x,y))
\end{gather} 

Several variants of FGSM have been proposed in recent literature with the goal of improving the effectiveness of the perturbation in increasing the loss function. 

\textbf{Random Step-FGSM (R-FGSM):} This method proposes to take a single random step followed by the application of FGSM. ~\cite{ens_adv_train} hypothesized that the loss function tends to be non-smooth near data points. Thus, taking a random step before applying FGSM should improve the quality of the gradient.

\textbf{Iterative-FGSM (I-FGSM):} Instead of taking just one step in the direction of the gradient, ~\cite{bim} proposed taking multiple smaller steps ($k$ steps of size $\epsilon/k$) with gradient being computed after each step.
\begin{gather} 
g_{t} = \nabla_{x} J(\theta,x'_{t},y) \\
x'_0 = x,  \ \ \ \ x'_{t+1} = x'_{t} + (\epsilon/k) \cdot sign(g_{t}) \label{x_update}
\end{gather}

\textbf{Momentum Iterative-FGSM (MI-FGSM):} ~\cite{momentum} observed that Iterative-FGSM can be improved to avoid poor local minima by considering the momentum information of gradient. They propose MI-FGSM which uses an exponential moving average of the gradient (i.e. momentum) to compute the direction of perturbation in each iteration.
\begin{gather} 
g_{t+1} = \mu \cdot g_{t} +  \frac{\nabla_{x} J(\theta,x'_{t},y)}{\rVert \nabla_{x} J(\theta,x'_{t},y)\rVert_{1}}
\end{gather} 
$\mu$ is the decay factor used to compute the moving average of the gradients. $x'_{t+1}$ is computed as shown in Eqn.\eqref{x_update}. This attack won the first place in the NIPS 2017 adversarial attack competition.

\textbf{PGD-CW:} This is a variant of the I-FGSM attack with the hinge loss function suggested by ~\cite{CarliniW16a}. Similar to ~\cite{madry2018towards}, we use Projected Gradient Descent (PGD) to maximize the loss function, which ensures that the attacked image lies within the $l_{\infty}$ ball around the natural image.


\ignore{
1. Goal: To reduce the transferability of adversarial examples and improve robustness to black-box attacks. We do this be reducing the adv subspace of the target model. Our insight is that we can use an ensemble of diverse models as explained in section 1.x. Our approach is a. find a way to measure the dimensionality of the shared adversarial subspace. 2. Use this measure during the training of ensembles to encourage diversity.
2. Measuring shared adversarial subspace: Prior works have tried to measure dim of adv subspace using methods like gradient aligned adv subspace. Unfortunately, this cant be used during backpropagation. We propose a way to differentiable way to measure adv dim so that it can be used in backprop. We explain our method using an example: (consider an ensemble of two models)--optimal direction of attack 
}

\section{Diversity Training}
\subsection{Approach}
The goal of our work is to improve the adversarial robustness of the model to black-box attacks by reducing the transferability of adversarial examples. This can be achieved by reducing the dimensionality of the Adv-SS of the target model using an ensemble of diverse models. Our approach to training an ensemble of diverse models is as follows:
\begin{enumerate}
\item Find a way to measure the dimensionality of Adv-SS of the ensemble
\item Use this measure as a regularization term to train an ensemble of diverse models.
\end{enumerate}
The rest of this section describes the two parts of our solution in greater detail.

\subsection{Measuring Adversarial Subspaces}
Adversarial examples have been known to span a contiguous subspace of the input. Several methods such as Gradient Aligned Adversarial Subspace ~\cite{space} and Local Intrinsic Dimensionality ~\cite{lid} have been proposed in recent literature to measure the dimensionality of this subspace. Unfortunately, these methods cannot be used in the cost function during training as computing them is either expensive or involves the use of non-differentiable functions. For our purposes, we want a computationally inexpensive way of measuring the dimensionality of Adv-SS using a differentiable function. This would allow us to run backpropagation through the function and use it as a regularization term during training.

\begin{figure*}[htb]
	\centering
    \centerline{\epsfig{file=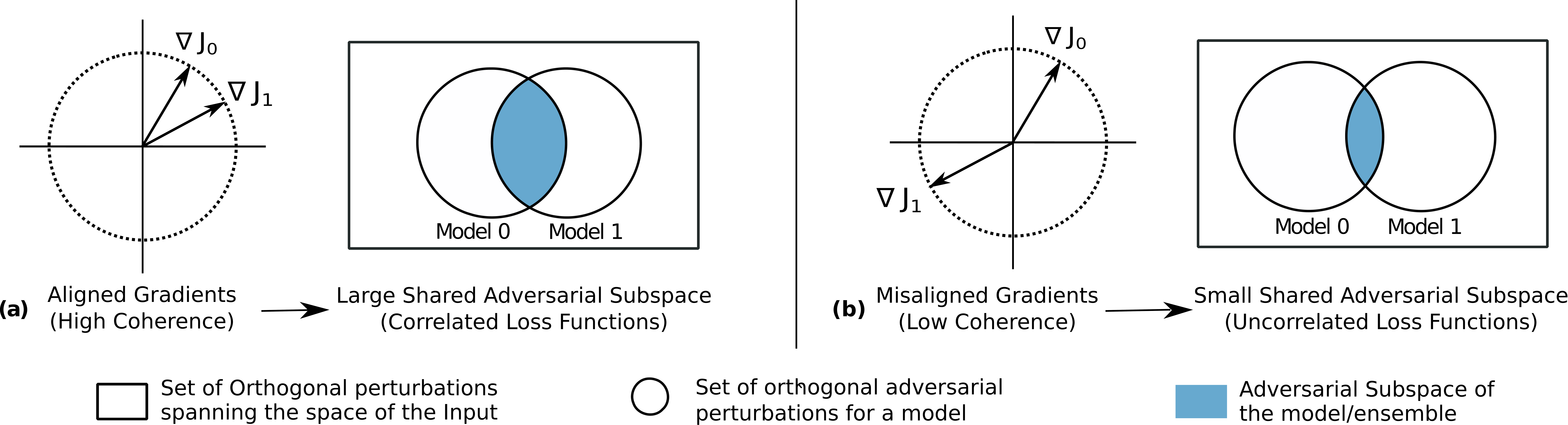, height=1.75in}}
	\caption{Illustration showing the relationship between the gradient alignment and overlap of adversarial subspaces of two models. Misaligned gradients indicate a smaller overlap in the adversarial subspace. }
	\vspace{0.1 in}
    \label{fig:grad_align}
\end{figure*}

\subsubsection{Adv-SS of an Ensemble}
Our proposal is to use an ensemble of models and thus we are interested in measuring the Adv-SS of the ensemble instead of a single model. For an input to be adversarial, it has to fool multiple models in the ensemble, requiring the example to lie in the shared Adv-SS of multiple models as shown in Figure~\ref{fig:motivation}b. Thus, the overall dimensionality of the Adv-SS of an ensemble is proportional to the amount of overlap in the Adv-SS of the individual models. We propose a novel method to measure this overlap by considering the alignment of the loss gradient vectors of the models.

\subsubsection{Gradient Alignment}
\ignore{
We first describe how the overlap of Adv-SS can be measured between two models and then show that our idea can be  generalized to an ensemble of n models.
-jacobians describe the direction of input perturbation that maximally increases the loss.
-intuitively, two models can be thought to have large overlap in adversarial subspaces, if similar perturbations fool both models. 
-Thus, alignment of jacobians can be used as a proxy to measure the overlap in the adversarial subspaces of two models.
-We 
}

We first describe how the overlap of Adv-SS can be measured between two models and then show that our idea can be  generalized to an ensemble of $N$ models. Consider two models $f_0$ and $f_1$. Let $\nabla_{x} J_{0}(\theta_{0},x,y)$ and $\nabla_{x} J_{1}(\theta_{1},x,y)$ denote the gradient of the loss functions of the two models with respect to the input $x$.

The gradient describes the direction in which the input $x$ has to be perturbed to maximally increase the loss function (locally around $x$). If the gradients of the two models are aligned, their loss functions increase in a correlated fashion, which means that a perturbation that causes $J_{0}$ to increase would likely also cause $J_{1}$ to increase. This indicates that the two models have similar adversarial directions and hence have a large shared Adv-SS as shown in Figure~\ref{fig:grad_align}a. In other words, adversarial examples that fool $f_{0}$ are also likely to fool $f_{1}$. Conversely, if the gradients of the two models are misaligned, the perturbations that cause $J_{0}$ to increase do not cause $J_{1}$ to increase. Thus $f_{0}$ and $f_{1}$ are unlikely to be fooled by the same perturbations implying that there is a reduction in the dimensionality of the shared Adv-SS of the two models as illustrated in Figure~\ref{fig:grad_align}b.

Thus, alignment of gradients can be used as a proxy to measure the amount of overlap in the Adv-SS of two models. A straightforward way to measure the alignment of gradients is to compute the cosine similarity (CS) between them.
\begin{gather} 
CS(\nabla_{x} J_{0}, \nabla_{x} J_{1}) = \frac{<\nabla_{x} J_{0}, \nabla_{x} J_{1}>}{|\nabla_{x} J_{0}| \cdot |\nabla_{x} J_{1}|}
\end{gather}

Cosine similarity has values in the range $[-1,+1]$. For two models, we would ideally like the cosine similarity to be -1 so that the gradients are completely misaligned and there is no overlap in the adversarial subspaces between the two models. For an ensemble of N models, we want the set of N gradient vectors $\{\nabla_{x} J_i\}_{i=1}^{N}$ to be maximally misaligned. 
One way of measuring the amount of alignment for a set of vectors is by considering their coherence value~\cite{coherence}. Coherence measures the maximum cosine similarity between unique pairs of vectors in the set.
We define coherence as shown in Eqn.~\ref{eq:coherence}.

\begin{gather} \label{eq:coherence}
coherence\big(\{\nabla_{x} J_i\}_{i=1}^{N}\big) = \max_{\substack{a,b \in \{1,..,N\} \\ a\neq b}} CS(\nabla_{x} J_{a}, \nabla_{x} J_{b})
\end{gather}

\ignore{
To measure the shared Adv-SS of an ensemble with $N$ models, we can consider the average cosine similarity ($CS_{avg}$) between each pair of models. 

\begin{gather} \label{eq:cs_avg}
CS_{avg} =\frac{1}{\binom{N}{2}} \ \ \ \ \sum_{\mathclap{1\leq a<b\leq N}}\ \ \ CS(\nabla_{x} J_{a}, \nabla_{x} J_{b})
\end{gather}
}

\ignore{
Cosine similarity has values in the range $[+1,-1]$. For two models, we would ideally like the cosine similarity to be -1 so that the gradients are completely misaligned and there is no overlap in the adversarial subspaces between the two models. For an ensemble of N models, we want the set of N gradient vectors  to be maximally misaligned. 
This is called the {\em coherence minimization problem}. We define coherence as shown in Eqn.~\ref{eq:coherence}
}

Coherence can be computed by taking the pairwise cosine similarity with vectors in $\{\nabla_{x} J_i\}_{i=1}^{N}$ and considering the $max$ over all the cosine similarity terms. Since this is a non-smooth function, minimizing Eqn.~\ref{eq:coherence} using first order methods like gradient descent would be slow. 
The rate of convergence can be improved by using a smooth approximation of this function ~\cite{Nesterov2005}. We replace $max$ in Eqn.~\ref{eq:coherence} with $LogSumExp$ as shown in Eqn.~\ref{eq:gal}. 
\begin{gather} \label{eq:gal}
GAL = \log\Bigg(\ \ \ \ \sum_{\mathclap{1\leq a<b\leq N}}\ \ \ \exp(CS(\nabla_{x} J_{a}, \nabla_{x} J_{b}))\Bigg)
\end{gather}

We call this term the {\em Gradient Alignment Loss (GAL)}. GAL can be used to approximate coherence and hence provides a way to measure the degree of overlap in the Adv-SS of the models in the ensemble.


\subsection{Diversity Training}
 If the models in an ensemble have a low GAL value for input $x$, it becomes harder to generate an adversarial example $x'$ that fool multiple models . We call such a collection of models with low GAL value a {\em diverse ensemble}. In order to encourage ensembles to have low GAL, 
 we propose using it as part of the cost function during training as a regularization term. We term this training procedure {\em Diversity Training (DivTrain)}.  Eqn.~\ref{eq:loss} shows the modified loss function. 

\begin{gather} \label{eq:loss}
Loss = \frac{1}{N} \sum_{i=1}^{N} J(\theta_i,x,y_{true}) + \lambda . GAL
\end{gather}
 
 The first term is the average cross entropy (CE) loss of each model in the ensemble and the second term represents the GAL. $\lambda$ is a hyperparameter that controls the importance given to GAL during training i.e. a large value of $\lambda$ would improve adversarial robustness at the cost of clean accuracy. 
 DivTrain lowers the dimensionality of the Adv-SS of the ensemble by reducing the overlap in the Adv-SS of the individual models. This reduces the transferability of adversarial examples and improves the adversarial robustness of the target model.

\subsection{Problem of Sparse Gradients}

\begin{figure}[htb]
	\centering
    \centerline{\epsfig{file=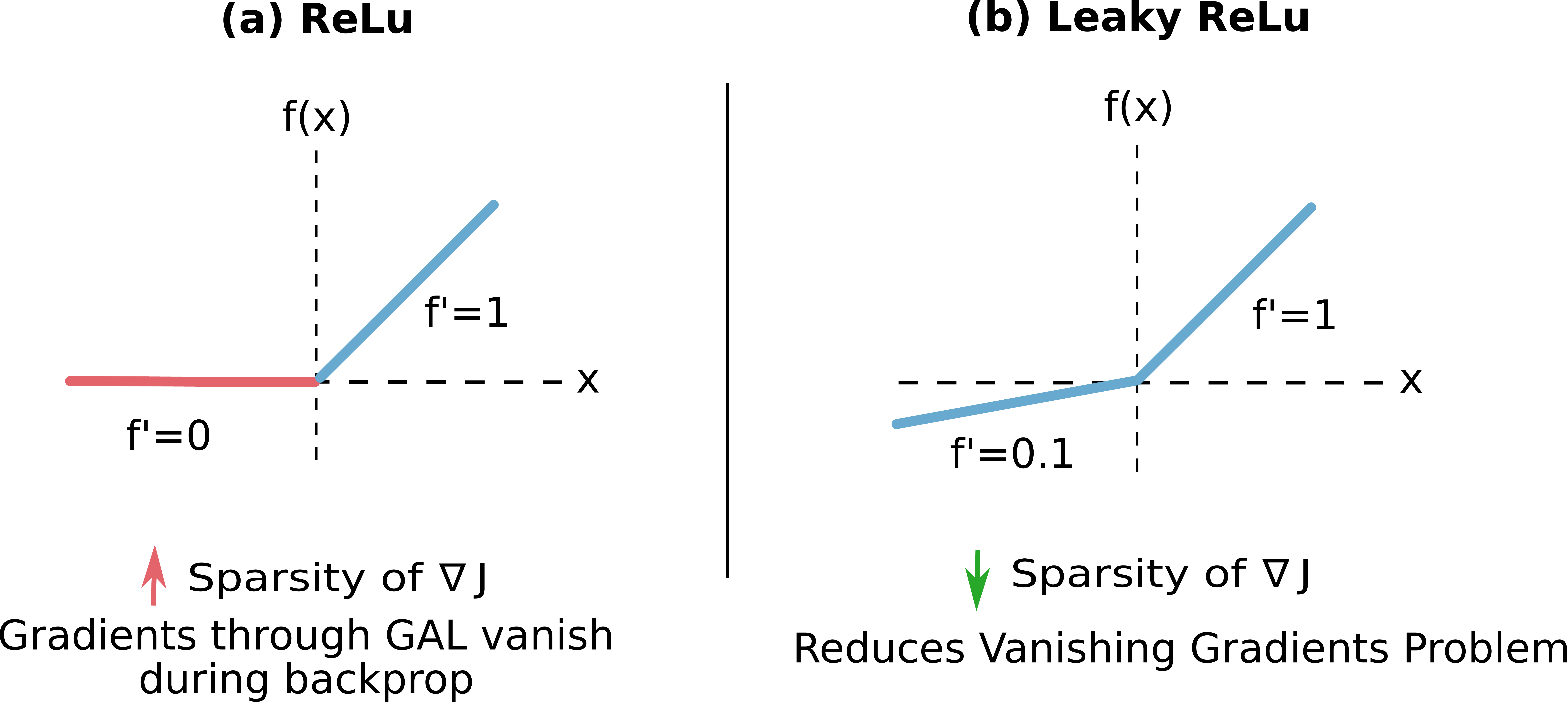, width=\columnwidth}}
	\caption{Leaky Relu improves backpropagation through GAL by reducing the sparsity of the Gradient Vector }
    \label{fig:relu}
\end{figure}

The selection of the activation function is an important design choice for the effective training of the network with GAL regularization. Using ReLu non-linearity in our networks, for example, causes the loss-gradient vector $\nabla_{x} J(\theta,x,y)$ to have a large number of zero values. This is because the derivative of the ReLu activation is zero valued in the saturating regime of the input ($x<0$) as shown in Figure~\ref{fig:relu}a. Since computing the GAL involves taking the inner product of $\nabla_{x} J$ terms, we end up with a large number of zero-valued product terms. This poses a problem for backpropagation since gradients don't flow through zero-valued products, causing the gradients to vanish and preventing the network from being trained. Our solution is to use an activation function which does not suffer from this problem like Leaky-ReLu instead. Leaky-Relu (Figure~\ref{fig:relu}b) has the desirable property of having a non zero gradient value regardless of the value of the input. This reduces the sparsity of the loss-gradient vector and improves the effectiveness of GAL regularization.

\ignore{
Experiments
-----------
-we conduct experimetns on mnist and cifar-10 datasets.
-We consider an ensemble trained with cross entropy loss and one trained with diversity training and show that diversity training reduces transferrability of adversarial exampels and improves robustness to bbox attacks . Sec 4.1 describes our experimental setup, 4.2 shows the results

1.Setup
We evaluate robustness of the target model under bbox attack conditions
-ADv does not have access to target model, so a surrogate is condiered.
-Our method is applicable to ensembles, so we compare two ensembles each with 5 identical models. One trained with just cross entropy loss and the other with diversity training.
}

\newcommand{\Base}{Base}
\newcommand{\Div}{Div}
\newcommand{\Ens}{Ens}
\newcommand{\EnsDiv}{Ens+Div}
\newcommand{\EAT}{EnsAdvTrain }
\newcommand{\DT}{DivTrain }

\section{Experiments}
We conduct experiments using the MNIST and CIFAR-10 datasets to validate our claims of improved robustness to transfer-based black-box attacks with DivTrain. We start by describing our experimental setup in Section~\ref{sec:setup}. Experimental results, including evaluations of combining DivTrain with the existing state of the art black-box defense is presented in Section~\ref{sec:results}. 
We compare the distributions of coherence values between diverse and baseline ensembles in Section~\ref{sec:cs_hist}.
Finally, we provide results for the Gradient Aligned Adversarial Subspace analysis in Section~\ref{sec:gaas} and show that \DT reduces the dimensionality of adversarial subspace of the ensemble.

\subsection{Setup}\label{sec:setup}
Our experiments evaluate the robustness of a Target model $T$ to black-box attacks. The attack is carried out by conducting white-box attacks on a Surrogate model $S$ and checking the accuracy of the adversarial examples generated in this way on $T$. We assume that the adversary has knowledge of the network architecture and the dataset used to train $T$ but not its model parameters. Hence, in our evaluations, we use models trained separately on the same dataset and having the same network architecture for $T$ and $S$. 
We evaluate the effectiveness of DivTrain by comparing the adversarial robustness between a baseline ensemble ($T_{\Base}$) trained with CE loss, without any regularization and a Diverse Ensemble ($T_{\Div}$) trained with GAL regularization. 
An ensemble with 5 models is used for both the target and surrogate model. We found that using an ensemble as a surrogate produced more transferable adversarial examples as compared to a single model. This is in line with the observations made by ~\cite{momentum,transferable_adv_examples}.

\begin{table}[h!]
\caption{Structure of the models used in our evaluations}
\label{table:config}
\vskip 0.15in
\begin{center}
\begin{small}
\begin{sc}

\begin{tabularx}{\columnwidth}{>{\hsize=0.22\hsize}X>{\hsize=0.78\hsize}X}
\toprule
 Model &Structure\\
\midrule
 Conv-3     & C32-C64-M-C128-M-FC1024-FC10\\
 Conv-4     & C32-C64-C128-M-C128-M-FC1024-FC10\\
 ResNet-20  & C16 - 3x\{RES16-RES32-RES64\} - FC10\\
 ResNet-26  & C16 - 4x\{RES16-RES32-RES64\} - FC10\\
\bottomrule
\end{tabularx}
\end{sc}
\end{small}
\end{center}
\vskip -0.1in
\end{table}

\newcommand\T{\rule{0pt}{2.6ex}}       
\newcommand\B{\rule[-1.2ex]{0pt}{0pt}} 

\begin{table*}[tb]
\caption{Classification accuracy of models under various black-box attacks comparing 1. Baseline-Ensemble ($T_{\Base}$) 2. Diverse-Ensemble ($T_{\Div}$) 3. \EAT ($T_{\Ens}$) and 4. \EAT + \DT defense ($T_{\EnsDiv}$). We use $\epsilon=0.1/0.3$ for MNIST and $\epsilon=0.03/0.09$ for CIFAR-10. The most successful attack for each perturbation size is highlighted in bold. $T_{\Div}$ improves adversarial robustness over $T_{\Base}$. The combined defense $T_{\EnsDiv}$ is more robust compared to the individual defenses ($T_{\Ens}$ / $T_{\Div}$).}
\label{table:results}
\vskip 0.15in
\begin{center}
\begin{small}
\begin{sc}
\begin{tabularx}{\textwidth}{p{1.5cm}p{1.5cm}p{1.0cm}XXXXX}
\toprule
 Model &Target(T)&Clean &FGSM &R-FGSM &I-FGSM &MI-FGSM &PGD-CW\\
\midrule
\multirow{4}{5em}{Conv-3 (mnist)} & $T_{\Base}$ & 99.4 & 91.4 / 9.7 & 92.0 / 9.7 & 86.1 / \textbf{0.7} & \textbf{85.7} / 2.6 & 92.3 / 9.7\\
  & $T_{\Div}$ & 99.2 & 97.1 / 34.3 & 97.8 / 30.6 & 97.6 / 20.4 & \textbf{96.9} / \textbf{16.9} & 97.1 / 35.9\\
  & $T_{\Ens}$ & 99.4 & 98.9 / 61.3 & 99.0 / \textbf{42.5} & 99.0 / 56.3 & \textbf{98.8} / 45.9 & 98.8 / 44.8\\
  & $T_{\EnsDiv}$ & 99.3 & 98.9 / 73.7 & 99.0 / 79.3 & 99.0 / 87.0 & 98.8 / \textbf{61.4} & \textbf{98.2} / 71.3\\ 
\hline \T
\multirow{4}{5em}{Conv-4 (cifar-10)} & $T_{\Base}$ & 85.1 & 14.1 / 7.8 & 16.8 / 3.2 & 9.5 \ \ / \textbf{2.8} & 9.0 \ \ / 7.4 & \textbf{8.8} \ \ / 5.9\\
  & $T_{\Div}$ & 82.4 & 45.3 / 14.7 & 56.0 / 15.1 & 51.4 / \textbf{5.6} & \textbf{35.0} / 7.5 & 43.9 / 11.6\\
  & $T_{\Ens}$ & 82.9 & 64.6 / 43.2 & 70.5 / 54.9 & 69.4 / 54.3 & \textbf{59.9} / \textbf{38.6} & 62.1 / 42.8\\
  & $T_{\EnsDiv}$ & 80.5 & 68.5 / \textbf{54.2} & 72.0 / 66.7 & 72.4 / 66.3 & \textbf{66.9} / 55.4 & 66.9 / 54.3\\
\hline \T
\multirow{4}{5em}{Resnet-20 (cifar-10)} & $T_{\Base}$ & 88.9 & 28.8 / 13.1 & 25.7 / 7.1 & \textbf{8.6} \ \ / \textbf{3.2}& 10.2 / 6.3 & 18.7 / 10.2\\
  & $T_{\Div}$ & 84.0 & 58.4 / 32.4 & 64.3 / 23.9 & 67.7 / 44.2 & \textbf{50.0} / \textbf{11.7} & 53.2 / 25.4\\
  & $T_{\Ens}$ & 87.9 & 70.9 / 44.3 & 77.2 / 50.9 & 79.6 / 65.5 & 66.5 / \textbf{30.5} & \textbf{65.9} / 37.8\\
  & $T_{\EnsDiv}$ & 84.7 & 74.9 / 50.7 & 78.1 / 57.6 & 79.7 / 71.5 & 74.1 / 47.4 & \textbf{71.3} / \textbf{46.3}\\
\hline \T
\multirow{4}{5em}{Mix (cifar-10)} & $T_{\Base}$ & 89.7 & 27.9 / 9.6 & 30.9 / 6.1 & 13.9 / \textbf{3.1} & \textbf{10.6} / 5.9 & 26.0 / 7.6\\
  & $T_{\Div}$ & 88.2 & 55.8 / 23.2 & 65.1 / 25.0 & 61.8 / 19.7 & \textbf{42.4} / \textbf{7.1} & 55.9 / 22.2\\
  & $T_{\Ens}$ & 87.4 & 72.6 / 49.6 & 76.9 / 58.4 & 77.1 / 61.4 & \textbf{66.9} / \textbf{27.9} & 70.0 / 47.1\\
  & $T_{\EnsDiv}$ & 86.4 & 73.4 / 52.2 & 77.9 / 64.1 & 77.2 / 67.5 & \textbf{69.2} / \textbf{39.7} & 71.8 / 50.4\\
\bottomrule
\end{tabularx}
\end{sc}
\end{small}
\end{center}
\vskip -0.1in
\end{table*}

\textbf{Network Configuration:} 
Table~\ref{table:config} lists the structure of the models used in our experiments. We use neural networks consisting of Convolutional(C), Max-pooling (M) and Fully Connected (FC) layers. The {\em Conv} networks have $3\times3$ Convolutional layers and $2\times2$ max-pooling layers. The {\em ResNet} structure is similar to the one described in ~\cite{resnet}. The network has total of $6n+2$ layers consisting of Residual blocks ({\em RES}) with skip connections after every two layers. 
Leaky-Relu is used as the non-linearity after each convolutional layer for all the networks.

\ignore{
 Additionally we augment both datasets with images perturbed by Gaussian noise: $\mathcal{N}(0,\,\epsilon)$. We care about loss functions of the models being uncorrelated primarily around natural images so that perturbations to the input do not cause loss functions of multiple models to increase. Augmenting the dataset with gaussian noise allows us to sample from a distribution that covers our region of interest in the input space.
}

The models are trained using the Adam optimizer~\cite{adam} with a learning rate of $0.001$. 
From our training dataset $\mathcal{D}$, we dynamically generate a augmented dataset $\mathcal{D'}$  using random shifts and crops for MNIST and random shifts, flips and crops for CIFAR-10.
In addition, we generate a dataset $\mathcal{D}_{noise}$ by adding a perturbation drawn from a truncated normal distribution: $\mathcal{N}(\mu=0,\sigma=\epsilon/2)$ to the images in $\mathcal{D}$ with $\epsilon=0.3$ for MNIST and $\epsilon=0.09$ for CIFAR-10. The combined dataset $\{\mathcal{D'},\mathcal{D}_{noise}\}$ is used to train diverse ensembles.
We care about reducing the coherence of $\{\nabla_{x} J_i\}_{i=1}^{N}$ primarily around natural images. Adding Gaussian noise to the training images allows us to sample from a distribution that covers our region of interest in the input space.

Conv-3 is trained for 10 epochs on MNIST dataset. Conv-4 and Resnet-20 are trained for 20 and 40 epochs respectively on CIFAR-10.
In addition, we also evaluate our defense with an ensemble consisting of mixture of different model architectures. The MIX ensemble is made up of \{Resnet-26, Resnet-20, Conv-4, Conv-3, Conv-3\} and is trained on CIFAR-10.
We set $\lambda = 0.5$ for DivTrain (Equation~\ref{eq:loss}) as we empirically found this to offer  a good trade-off between clean accuracy and adversarial robustness.

\textbf{Attack Configuration:} 
We evaluate the accuracy of the target models $(T_{Base}, T_{Div})$ for clean examples as well as adversarial examples crafted from black-box attacks using the attacks described in Section~\ref{ssec:wb_attacks} on Surrogate $S$. Since $S$ is an ensemble of 5 models, we consider the average CE loss of all the models in $S$ as the objective function to be maximized for the attacks. We briefly describe the parameters for each attack: FGSM takes a single step of magnitude $\epsilon$ in the direction of the gradient. R-FGSM applies FGSM after taking a single random step sampled from a uniform distribution: $\mathcal{U}(-\epsilon,+\epsilon)$. For I-FGSM and MI-FGSM, we use $k=10$ steps with each step of size $\epsilon/10$. The decay factor $\mu$ is set to $1.0$ for MI-FGSM. We evaluate PGD-CW with a confidence parameter $k=50$ and $30$ iterations of optimization considering the hinge-loss function from ~\cite{CarliniW16a}. Results are reported for two different perturbation sizes for all the attacks: $\epsilon = 0.1/0.3$ for MNIST and $\epsilon=0.03/0.09$ for CIFAR-10. We use the Cleverhans library's~\cite{cleverhans} implementation of the attacks to evaluate our defense.

\begin{figure*}[htb]
	\centering
    \centerline{\epsfig{file=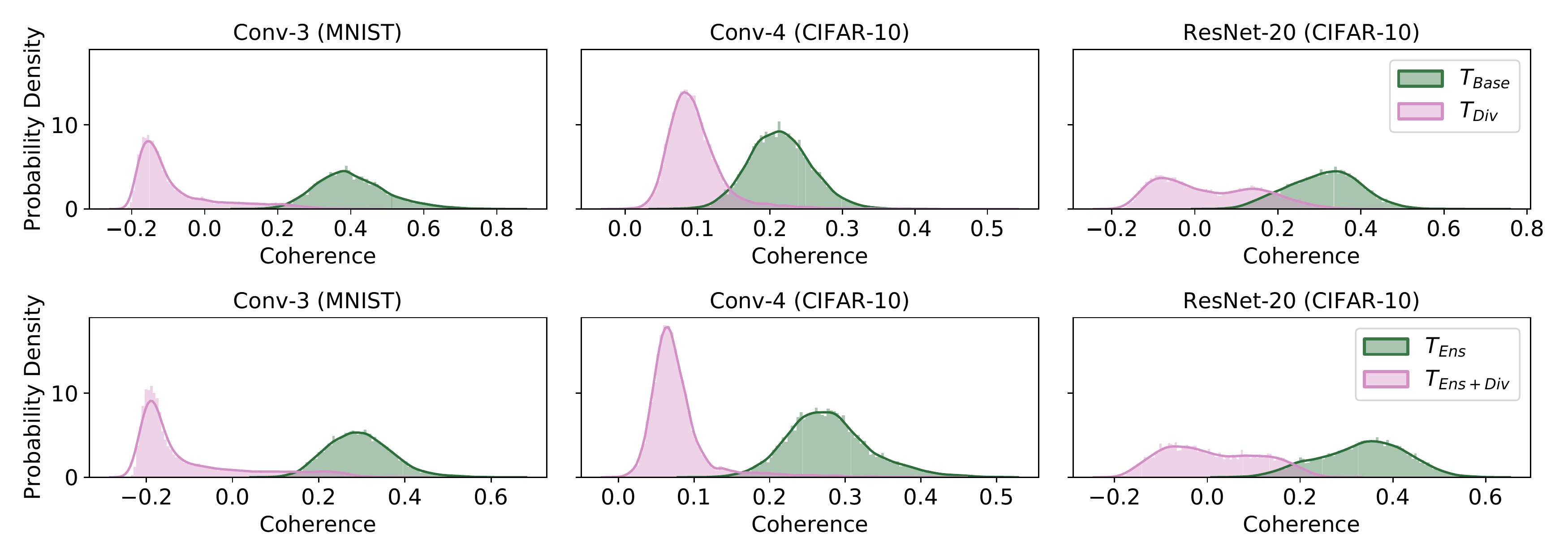, height=2.4in}}
	\caption{Histogram of Coherence values plotted for Conv-3, Conv-4 and Resnet-20 comparing $T_{\Base}$, $T_{\Div}$, $T_{\Ens}$ and $T_{\EnsDiv}$. Models trained with Diversity Training (GAL regularization) have misaligned gradient vectors with lower coherence values.}
	\vspace{0.1 in}
    \label{fig:coherence}
\end{figure*}

\subsection{Results}\label{sec:results}
\ignore{
Tbase, Tdiv
comment on which attacks are better

}
Our results comparing Baseline-Ensemble ($T_{\Base}$) and Diverse-Ensemble ($T_{\Div}$) are shown in Table~\ref{table:results} with the most effective attack highlighted in bold. $T_{\Div}$ has a significantly higher classification accuracy on adversarial examples compared to $T_{\Base}$ for all the attacks considered, showing that DivTrain improves the adversarial robustness of ensembles against black-box attacks. The classification accuracy of clean examples drops slightly as a consequence of adding the regularization term (GAL) to the cost function. 

There are several defenses proposed in recent literature to defend against black-box attacks. Since DivTrain is a defense that is generally applicable to any ensemble of models,  
it can be combined with existing proposals to create a stronger defense. We show that this is possible by evaluating a combination of our method with Ensemble Adversarial Training~\cite{ens_adv_train} , which is the current state of the art method for Black box defense.

\textbf{Combined Defense:} Ensemble Adversarial Training (\EAT) proposes to improve adversarial robustness by augmenting the training dataset with adversarial examples generated from a static pre-trained model. We use a pre-trained model with the same architecture as the target model. By attacking the pre-trained model with FGSM, we generate an adversarial dataset $\mathcal{D}_{adv}$. The target models are trained on the combined dataset consisting of both clean and adversarial examples $\{\mathcal{D'},\mathcal{D}_{adv}\}$. Perturbation sizes used to generate the adversarial examples are randomly determined from a truncated normal distribution $\mathcal{N}(\mu=0,\sigma=\epsilon/2)$ to ensure adversarial robustness against various perturbation sizes as suggested by ~\cite{bim}.

We denote the ensemble trained with EnsAdvTrain as $T_{\Ens}$ and the ensemble trained with the combination of \EAT and \DT as $T_{\EnsDiv}$. Results showing the classification accuracy of $T_{\Ens}$ and $T_{\EnsDiv}$  under various attacks are provided in Table~\ref{table:results}. The combined defense offers higher classification accuracy under attack compared with either of the two defenses $T_{\Ens}$/$T_{\Div}$ used alone. This shows that \DT can be combined with existing methods such as \EAT to create a stronger defense.

\subsection{Distribution of Coherence}~\label{sec:cs_hist}
Diversity Training encourages models to have uncorrelated loss functions by reducing the coherence among their gradient vectors. Figure~\ref{fig:coherence} compares the distribution of coherence values (see Eqn.~\ref{eq:coherence}) for the different target models used in our evaluations.  The histograms show that $T_{\Div}$ and $T_{\EnsDiv}$ have lower coherence values compared to $T_{\Base}$ and $T_{\Ens}$. Thus, our proposed GAL regularization is an effective way of training models with misaligned gradient vectors which can be used to create ensembles with improved adversarial robustness to black-box attacks.

\subsection{Gradient Aligned Adversarial Subspace}~\label{sec:gaas}
\begin{figure*}[htb]
	\centering
    \centerline{\epsfig{file=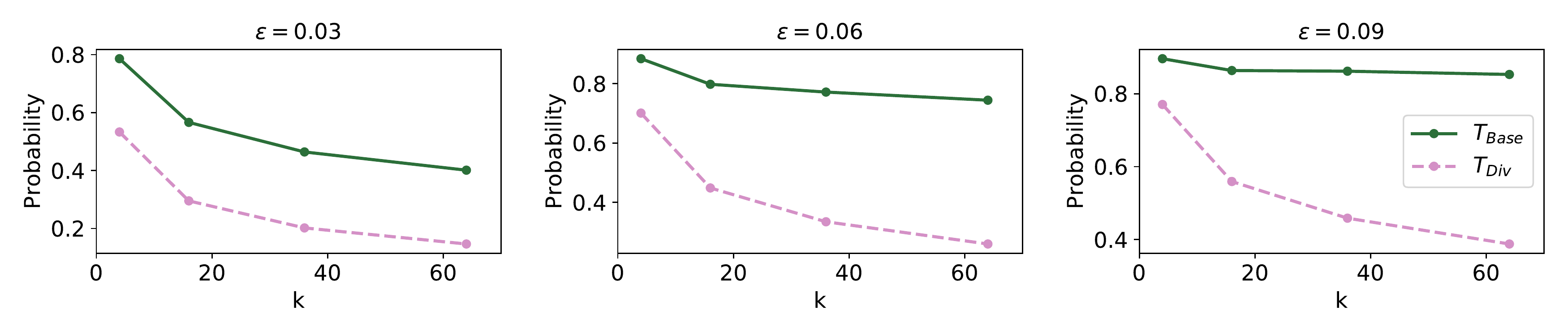, height=1.4in}}
	\caption{Gradient aligned adversarial subspace analysis performed on Conv-4 with CIFAR-10. Plots show the probability of finding $k$ orthogonal adversarial directions for 3 different perturbation sizes $\epsilon=0.03/0.06/0.09$. $T_{\Div}$ has consistently lower probabilities of finding an adversarial direction compared to $T_{\Base}$ showing that \DT lowers the dimensionality of the Adv-SS of an ensemble.}
	\vspace{0.1 in}
    \label{fig:gaas}
\end{figure*}

We provide further evidence that DivTrain lowers the dimensionality of the Adv-SS of the ensemble by using the Gradient Aligned Adversarial Subspace (GAAS) ~\cite{space} analysis. GAAS measures the dimensionality of the Adv-SS by aligning a set of $k$ orthogonal vectors $\{r_{i}\}_{i=1}^{k}$ with the gradient of the loss function $\nabla_{x} J(\theta,x,y)$ in order to find a maximal set of orthogonal adversarial perturbations. The orthogonal vectors are constructed by multiplying the row vectors from a Regular Hadamard matrix component-wise with $sign(\nabla_{x} J)$. For a Regular Hadamard matrix of order $k$, this yields a set of $k$ orthogonal vectors aligned with the gradient. We run the GAAS analysis on the Conv-4 model and use $k\in\{4,25,36,64\}$.  Figure~\ref{fig:gaas} compares the probability of finding successful orthogonal directions for $T_{\Base}$ and $T_{\Div}$. We repeat the analysis with different perturbation sizes ($\epsilon=0.03/0.06/0.09$). Our results show that $T_{Div}$ has fewer orthogonal adversarial directions compared to $T_{\Base}$ showing that \DT can effectively lower the dimensionality of the Adv-SS of ensembles.




\section{Related Work}
\ignore{
-Several defenses have been proposed for both white-box and black-box attacks(cite the works)
-Several white-box defenses suffer from gradient masking (defensive-distillation)
-As a result, these defenses have been shown to fail for more powerful attacks that take into account the defense mechanism
-also subject to transfer based attacks from a surrogate model that does not hide the gradient
-defenses have also been proposed for black box attacks both in the context of query access and no-query access. We discuss the ones that are closely related to our work.
-Defenses involving ensemble have been proposed with a similar idea as ours of improving adversarial robustness with diversity
-[Ensemble Methods as a Defense:] uses an ensemble of models without any special regularization. This relies on different initialization of the model/different model architectures to improve adversarial robustness and is equivalent to the Baseline ensemble we consider in our work. While ensembles do improve adversarial robustness compared to a single model, the robustness of ensembles to attacks can be greatly improved with diversity Training

-[Interpreting Adversarial Robustness: A View from Decision Surface in Input Space] considers the relationship between adversarial robustness and smoothness of the decision surface in the input space. This work proposes a robust training scheme involving the use of jacobian terms. The key insight is to make the jacobian values close to 0. We instead take the gradient values and try to encourage negative correlation in the context of an ensemble.

-

}
The susceptibility of deep neural networks to adversarial inputs has sparked considerable research interest in finding ways to make deep learning models robust to adversarial attacks. As a result, a number of methods have been proposed to defend against white-box attacks ~\cite{defensive_distillation,exp_harness,cascade,thermometer,input_transformations,lid,sap,randomization,pixeldefend,defense_gan}. However, a fair majority of these defenses have been shown to be ineffective against adaptive attacks that are tailor-made to work well against specific defenses. 
A recent work ~\cite{obfs_grad} showed that a number of these attacks rely on some form of gradient masking~\cite{practical_bbox}, whereby the defense makes the gradient information unavailable to the attacker, making it harder to craft adversarial examples. It also proposes techniques to tackle the problem of obfuscated gradients that break defenses which use gradient masking.

\ignore{
Adversarial Training ~\cite{exp_harness} is a defense that aims to improve adversarial robustness by including adversarial examples during training. ~\cite{ens_adv_train} showed that Adversarial Training with adversarial examples derived from the model being trained causes the loss function to converge to a degenerate global minima. It produces curvature artifacts near the data points which has the effect of masking the true gradient of the loss function making the attack ineffective.
To remedy this, they propose Ensemble Adversarial Training, which involves using adversarial examples generated from an alternate model to train the target model. This was shown to improve the adversarial robustness of the model to transfer-based black-box attacks. We use this defense in our experiments to show that diversity training works when combined with existing defenses.
}

Prior works have considered the use of ensembles to improve adversarial robustness.
~\cite{ensemble_defense} use ensembles with the intuition that adversarial examples that lead to misclassification in one model may not fool other models in the ensemble.
~\cite{random_self_ensemble} use noise injection to the layers of the neural network to prevent gradient based attacks and ensembles predictions over random noises. While both of these works benefit from the adversarial robustness offered by ensembles as discussed in Section~\ref{sec:ens_effective_defense}, the contribution of our work is to further improve this robustness by explicitly encouraging the models in the ensemble to have uncorrelated loss functions.

The idea of using cosine similarity of gradients to measure the correlation between loss functions has been explored in ~\cite{grad_similarity}. This was used in the context of improving the usefulness of auxiliary tasks to improve data efficiency. In contrast, we develop a metric based on cosine similarity that can measure  the similarity among a set of gradient vectors with the objective of measuring the overlap of adversarial subspaces between the models in an ensemble.
\ignore{
1. GAL to detect attacks
2. Adaptive Attacks
3. 
}

\section{Discussion}
Our paper explores the use of diverse ensembles with uncorrelated loss functions to better defend against transfer-based attacks. We briefly discuss other problem settings where our idea can potentially be used.


\textbf{Adversarial Attack Detection:} 
The objective here is to detect inputs that have been adversarially tampered with and flag such inputs. 
A recent work ~\cite{ensembles_detect} uses ensembles to detect adversarial inputs by training them to have high disagreement for inputs outside the training distribution. Since GAL minimizes the coherence of gradient vectors, it can potentially be used for the same purpose.
One possible approach would be to use a modified version of \DT which uses GAL as the cost function (without cross-entropy loss) for examples outside the training distribution so that the consensus among the members of the ensemble would be low for out-of-distribution data. 

\textbf{Better Black-Box Attacks:} Ensembles can also be used to generate better black-box attacks. ~\cite{momentum, transferable_adv_examples} use an ensemble of models as the surrogate with the intuition that adversarial examples that fool multiple models tend to be more transferable. It would be interesting to study the transferability of adversarial examples generated on diverse-ensembles to see if they can enable better black-box attacks. We leave the evaluation of both of these ideas for a future work.


\ignore{
-Transfer based attacks
-Ensembles can be used to reduce dim of advss - harder to fool multiple models
-Diversity training can reduce the overlap of advss and help improve robustness
-We show empirically that models trained with div train are less susceptible to transfer based attacks
-Further, we show that div train can be combined with existing defense like ensemble adv train to further improve adv robustness
-Our claims of reduced dim of advss are validated using the GAAS analysis
-We also provide the distribution of pair-wise cosine similarity of div-trained ensembles and show that GAL regularization effectively reduces correlation between the loss functions of the ensemble by penalizing cosine similarity.
}

\ignore{
-intuition: Harder to fool multiple models if they have uncorrelated loss functions
-Ensembles with uncorrelated loss functions can be used to better defend against transfer based attacks. 
-Reducing correlation between loss functions is equivalent to minimizing the coherence of the gradient vectors to reduce the amount of overlap between adversarial subspaces.
-We develop a training method called Diversity Training to train an ensemble of diverse models.
-We propose a novel regularization term called GAL and show that it can be used to minimize coherence to train a diverse model with improved adversarial robustness.
-Additionally, we show that DivTrain can be combined with an existing defense (Ensemble Adversarial Training) to further improve adversarial robustness of the ensemble.
We believe that reducing the dimensionality of adversarial subspace is important to achieving a robust defense and our work is a step in this direction.
}

\section{Conclusion}
Transfer-based attacks present an important challenge for the secure deployment of deep neural networks in real world applications. 
We explore the use of ensembles to defend against this class of attacks with the intuition that it is harder to fool multiple models in an ensemble if they have uncorrelated loss functions.
We propose a novel regularization term called {\em Gradient Alignment Loss} that helps us train models with uncorrelated loss function by minimizing the coherence among their gradient vectors. We show that this can be used to train a {\em diverse ensemble} with improved adversarial robustness by reducing the amount of overlap in the shared adversarial subspace of the models. Furthermore, our proposal can be combined with existing methods to create a stronger defense. We believe that reducing the dimensionality of the adversarial subspace is important to creating a strong defense and {\em Diversity Training} is 
a technique that can help achieve this goal.
\section*{Acknowledgements}
We thank our colleagues from the Memory Systems Lab for their feedback. This work was supported by a gift from Intel. We also gratefully acknowledge the support of NVIDIA Corporation with the donation of the Titan V GPU used for this research.

\bibliography{references}
\bibliographystyle{icml2019}

\end{document}